\documentclass[a4paper, 10pt, conference]{ieeeconf}      
\usepackage{FG2021}

\FGfinalcopy 

\usepackage{graphics} 
\usepackage{graphicx}
\usepackage{multirow}

\def\FGPaperID{272} 

\title{\LARGE \bf
GROWL: Group Detection With Link Prediction
}

\author{\parbox{16cm}{\centering
    {\large Viktor Schmuck and Oya Celiktutan}\\
    {\normalsize
    Centre for Robotics Research, Department of Engineering, King's College London, London, United Kingdom\\}}%
}

\begin{document}

\IEEEoverridecommandlockouts\pubid{\makebox[\columnwidth]{978-1-6654-3176-7/21/\$31.00~\copyright{}2021 IEEE \hfill}
\hspace{\columnsep}\makebox[\columnwidth]{ }}

\ifFGfinal
\thispagestyle{empty}
\pagestyle{empty}
\else
\author{Viktor Schmuck and Oya Celiktutan\\ 272 \FGPaperID \\}
\pagestyle{plain}
\fi
\maketitle

\begin{abstract}
Interaction group detection has been previously addressed with bottom-up approaches which relied on the position and orientation information of individuals. These approaches were primarily based on pairwise affinity matrices and were limited to static, third-person views. This problem can greatly benefit from a holistic approach based on Graph Neural Networks (GNNs) beyond pairwise relationships, due to the inherent spatial configuration that exists between individuals who form interaction groups. Our proposed method, GROup detection With Link prediction (GROWL), demonstrates the effectiveness of a GNN based approach. GROWL predicts the link between two individuals by generating a feature embedding based on their neighbourhood in the graph and determines whether they are connected with a shallow binary classification method such as Multi-layer Perceptrons (MLPs). We test our method against other state-of-the-art group detection approaches on both a third-person view dataset and a robocentric (i.e., egocentric) dataset. In addition, 
we propose a multimodal approach based on RGB and depth data to calculate a representation GROWL can utilise as input. Our results show that a GNN based approach can significantly improve accuracy across different camera views, i.e., third-person and egocentric views. 
\end{abstract}

\section{INTRODUCTION}
Robots are becoming more commonplace in our everyday lives. To be deployed in dynamic human environments, they need to carry out their tasks automatically and in a human-aware manner. For this, they must analyse crowded social scenes to identify people and understand how they can be grouped into interaction groups. Such analysis is essential to equip robots with the capability of navigating a crowded environment, such as a museum, airport or university, and approaching individuals or groups in a socially-aware manner to offer guidance or other help.

Previous research indicates that good results can be achieved via the identification of F-formations~\cite{kendon_conducting_1990}, which entails the detection of interaction spaces of groups of people. Most of the existing approaches for F-formation detection rely on images from static, third-person view recordings~\cite{bazzani_decentralized_2012, chandran_identifying_2015, chen_anchor-based_2017, elassal_unsupervised_2016, ge_vision-based_2012,  hedayati_reform_2020, hedayati_recognizing_2019, mazzon_detection_2013,  ricci_uncovering_2015, setti_f-formation_2015}. These approaches are not directly applicable to mobile robots, due to noise factors introduced by retrieving information from a robocentric view. Approaches targeting ego- and robocentric input are mainly focused on unsupervised methods~\cite{taylor_robot-centric_2020, schmuck_robocentric_2020}. However, using the structure of the holistic spatial layout of F-formations has not been investigated by the aforementioned works to achieve \textit{interaction group} detection.

\begin{figure}[t]
\centering
\tiny
\setlength\tabcolsep{0pt}
    \begin{tabular}{cc}
        \includegraphics[width=0.2\textwidth]{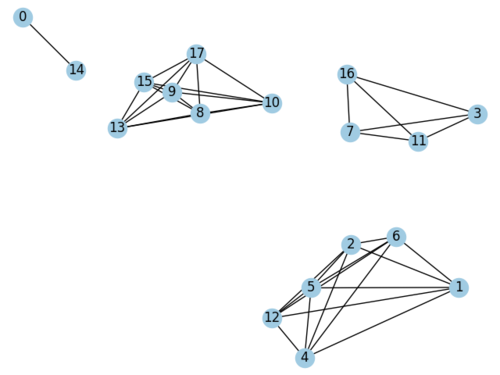} & 
        \includegraphics[width=0.2\textwidth]{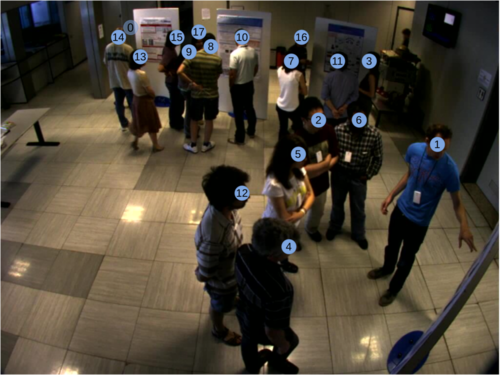}
        \\
      (a) & (b)
      \\
    \end{tabular}
    \caption{(a) Top-down representation of interaction groups; (b) Third-person view image from the SALSA Poster Session~\cite{alameda-pineda_salsa_2016} dataset. Given an image such as (b), the goal of our proposed approach is to produce the top-down representation in (a) from a third-person or egocentric view by first translating people's positions to a 2D, top-down view plane and then predicting who belong to the same interaction group. 
    }
    \vspace{-0.5cm}
    \label{fig:intro_detection}
\end{figure}

Graphs are a well-suited means for representing and identifying interaction groups due to the nature of the problem. Graphs have indeed been investigated before by a number of works~\cite{choi_discovering_2014, setti_f-formation_2015, zhang_beyond_2016}, but the usage of Graph Neural Networks (GNNs) has not been explored for this task. Only Yang et al.~\cite{yang_group_2020} used GNNs for 
analysing changes in group behaviour when a new person joins. While recent advances in GNNs in representation learning~\cite{hamilton_inductive_2018} and link prediction~\cite{zhang_link_2018} showed superior results for other tasks such as recommender systems, exploration of these techniques for interaction group detection has remained partial.

We introduce a novel approach, GROWL, to detect interaction groups (see Figure~\ref{fig:intro_detection}) by leveraging the power of GNNs and feature embedding on graphs. Given a social scene image, GROWL can use a graph based representation created from position and orientation data to generate feature embeddings for individuals. For robocentric data, we propose a method for computing the required graph based representation based on multimodal (RGB and depth image) data. These embeddings can be used as input to a binary classifier to determine whether two individiuals belong to the same interaction group. GROWL sets the state-of-the-art performance on both third-person view datasets such as the SALSA dataset~\cite{alameda-pineda_salsa_2016} and its subsets as well as robocentric datasets such as the RICA dataset~\cite{schmuck_rica_2020}. Our results show that 1) on the SALSA Poster Session dataset GROWL outperforms REFORM~\cite{hedayati_reform_2020}, the latest state-of-the-art interaction group detection method, by a margin of $10.2\%$; and 2) on the RICA dataset it outperforms a baseline approach, Graph-cuts (GCFF)~\cite{setti_f-formation_2015} by a margin of $20.8\%$ in terms of $F_1$-score. 

In summary, we have three central contributions: (1) We propose a novel approach to the problem of detecting interaction groups in social scenes based on a GNN that combines representation learning~\cite{hamilton_inductive_2018} and link prediction~\cite{zhang_link_2018}. (2) We propose a method based on multimodal data to represent detected individuals captured from a robocentric view in a top-down graph representation. (3) We show that our proposed approach can be  successfully applied to both third-person and egocentric view datasets. GROWL significantly improves group detection accuracy, outperforming a baseline method and the most related previous work. 
 
\section{RELATED WORK}
\label{sec:relwork}
In this section, we summarise the previous work on F-formation detection from images. F-formations were first defined by Kendon~\cite{kendon_conducting_1990} who used them to describe how a group of people arrange themselves when they interact with each other. This concept has been used in most previous work addressing interaction group detection. We consider an interaction group, based on the definition of F-formations, as two or more people in close proximity who engage in a common activity (e.g. see Figure~\ref{fig:intro_detection}).

To identify F-formations, Cristani et al.~\cite{marco_cristani_social_2011} proposed Hough Voting for F-formations (HVFF) which took positions and head orientations of identified individuals in third-person view scenes as inputs. Based on these features, they employed Hough Voting to identify F-formations and assign individuals to them. Setti et al.~\cite{setti_f-formation_2015} proposed a multi-scale extension of Hough Voting based on graph cuts (GCFF). In addition, they defined how the $F_1$-score based accuracy of group detection methods should be measured, which was adopted in this paper.

Vascon et al.~\cite{vascon_game-theoretic_2014} proposed an approach based on the position and orientation of people, which embedded constraints defined by F-formations into a game-theoretic probabilistic approach. They calculated attention frustums of people and modelled their pairwise relations before using the generated matrix in a non-cooperative game to produce clusters. Inaba et al.~\cite{inaba_conversational_2016} also proposed the calculation of affinity matrices; however, instead of detecting F-formations, they treated groups as dominant sets. To detect interaction groups, they defined a quadratic optimisation problem that they solved with a method also taken from game theory. Zhang and Hung~\cite{zhang_beyond_2016, zhang_social_2018} proposed a similar solution but instead of working with pairwise affinity matrices only, they modelled social involvement of individuals. This was done by calculating the relation of individuals' features to those calculated from entire groups they might be associated with.

Hedayati et al.~\cite{hedayati_reform_2020, hedayati_recognizing_2019} proposed an approach called REFORM which was similar to the pairwise affinity matrix based ones mentioned above. However, their method treats the problem of F-formation detection as a binary classification problem, i.e., whether two people belong to the same group or not. Their proposed classifiers such as Weighted KNN, Bagged Tree and Logistic Regression and greedy reconstruction method that aggregates pairwise relations for entire scenes achieved the latest state-of-the-art accuracy and good generality across multiple benchmark datasets such as SALSA~\cite{alameda-pineda_salsa_2016} and Babble~\cite{hedayati_reform_2020}. Their solution is similar to the one proposed by Ramírez et al.~\cite{ramirez_modeling_2016} who performed link prediction based on proxemics in between pairs of individuals to calculate whether they belong to the same group. Yang et al.~\cite{yang_group_2020} proposed an in-part Graph Convolutional Network based approach to analyse both skeleton and group information to perform group behaviour recognition. Their algorithm used skeleton, position and orientation data to analyse the effects of a person joining an interaction group. However, group detection from individuals was outside the scope of their solution.

A few works focused on unsupervised group detection in dynamic, egocentric views, particularly captured via a mobile robot. Taylor et al.~\cite{taylor_robot-centric_2020} proposed a method based on agglomerative hierarchical clustering (AHC), which relied on the position and velocity of detected individuals. Similarly, Schmuck et al.~\cite{schmuck_rica_2020} proposed an AHC based solution, investigating different linkage options, and their algorithm utilised position and depth information instead of the position and orientation features that were used by the aforementioned approaches~\cite{marco_cristani_social_2011, hedayati_reform_2020, hedayati_recognizing_2019, setti_f-formation_2015, vascon_game-theoretic_2014, zhang_social_2018}.

Taken together, previous work addressed group detection and F-formation detection with position and orientation information of people. The majority of works construct different pairwise affinity maps and are bottom-up, i.e., starting from detected individuals and building groups from them. Moreover, most of the methods rely on third-person view datasets to train and evaluate their solutions. Our proposed method, GROWL, makes use of the common position and orientation features but through a GNN based link-prediction approach. Our solution works without high-level information (i.e., skeleton data) and utilises group information to detect interaction groups in contrast to the work of Yang et al.~\cite{yang_group_2020},  which investigates changes in group behaviour when a new person joins an existing group. Moreover, our proposed work investigates an approach which can leverage the strengths related to pairwise affinity maps. On a more global scale and to our knowledge, it is the first to be tested in both third-person view and egocentric view settings. 

\section{BACKGROUND}
\label{sec:background}
Graph Neural Networks (GNNs) have been widely used in a broad range of computer vision tasks ranging from activity recognition based on skeleton data~\cite{yan2018spatial}, processing facial landmarks~\cite{Antonakos_2015_CVPR} and object parsing in scenes~\cite{liang2016semantic}. The GNNs are constructed from nodes and edges connecting them, where both nodes and edges can have associated features. Typically, node features hold embedded information calculated based on their neighbourhood, defined by their connected edges. Edge features are commonly computed based on the relationship between two nodes, similarly to pairwise affinity fields. GNNs have been shown to be better than ordinary neural network based approaches in many computer vision tasks that inherit spatial structures, and 
therefore GNNs are a natural means to learn existing relationships between unstructured input features. 

\subsection{Representation learning on graphs}
\label{sec:background_embedding}
Representation learning on graphs allows models not only to gather information from single nodes' raw input features, but also to learn aggregated representations based on their neighbourhood - i.e., defining their embedded set of features based on raw features or learnt feature embeddings of other nodes which are connected by an edge. Hamilton et al.~\cite{hamilton_inductive_2018} proposed GraphSAGE, a recent state-of-the-art approach to calculate embeddings for graph node features. GraphSAGE generates embeddings for nodes by learning how to aggregate feature information based on their neighbourhood. This is achieved by employing standard machine learning techniques such as forward- and backpropagation and using stochastic gradient descent. Nodes typically have different numbers of neighbouring nodes and they are unordered, therefore an embedding algorithm is required to be invariant to the size (i.e., number of nodes) and permutation of its inputs. To address this issue, Hamilton et al.~\cite{hamilton_inductive_2018} investigated mean-, LSTM-, and pooling-based aggregators and have concluded that the mean aggregator, while being marginally less accurate than other methods, is more computationally efficient.

\subsection{Link prediction with GNNs}
\label{sec:background_link}
Link prediction is the process of determining whether two nodes in a graph should be connected by an edge. 
Zhang and Chen~\cite{zhang_link_2018} proposed to use link prediction for subgraph detection. Their approach achieved state-of-the-art results and was justified using not only entire graphs but also local subgraphs for learning link prediction models. Moreover, they proposed \textit{negative injection} for training their GNN. Negative injection, while taking existing (positive) edges as samples, also uses samples of generated node embeddings based on non-existent (negative) edges. This method achieves better generalisation performance as the GNN does not overfit to predict existing edges due to the lack of non-existent edges in the training set.

Considering the previous works, it is possible to employ local GraphSAGE~\cite{hamilton_inductive_2018} embeddings, as Zhang and Chen's work~\cite{zhang_link_2018} suggests entire network embeddings are unnecessary and computationally complex. Moreover, negative injection should be implemented for better generalisation. 
\begin{figure}[b]
  \centering
  \includegraphics[width=0.35\textwidth]{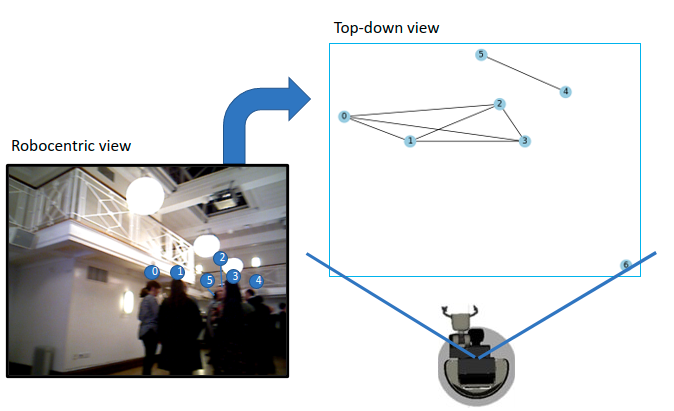}
  \caption{Transformation of a robocentric image (from the RICA~\cite{schmuck_rica_2020} dataset) into a top-down representation of people in the scene.
  Blue lines indicate the robot's field of view. Circles represent individuals and are mapped from the robocentric view to the top-down representation. Nodes connected by edges show which people form interaction groups. 
  }
  \vspace{-0.5cm}
  \label{fig:topdownrep}
\end{figure}

\begin{figure*}
\centering
\tiny
\setlength\tabcolsep{10pt}
    \begin{tabular}{cc}
        & \multirow{3}{*}[-10em]{\includegraphics[width=70mm]{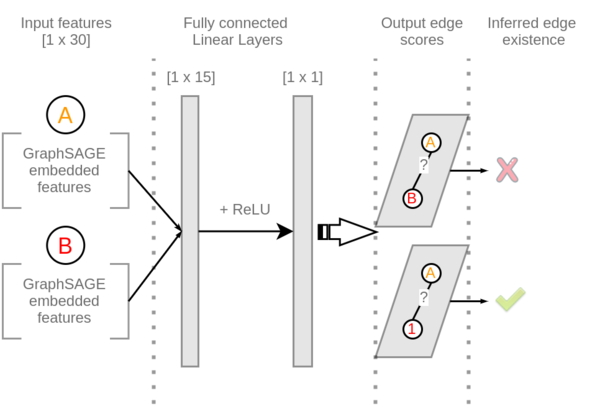}}
        \\
      \includegraphics[width=23mm]{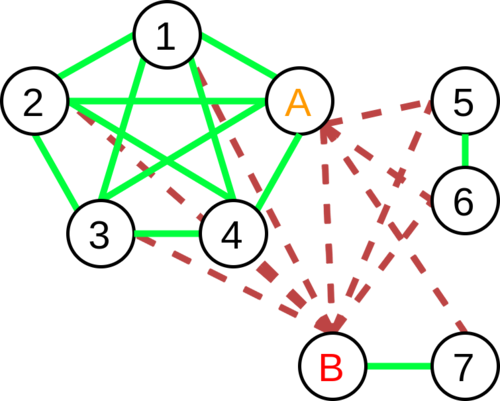} &
      \\
      (a) &
      \\ [0.5cm]
      \includegraphics[width=65mm]{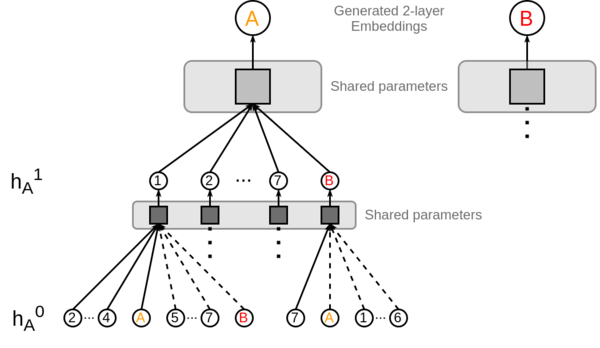} &
      \\
      (b) & (c)
      \\
    \end{tabular}
    \caption{(a) Fully connected graph used for training GROWL: green lines show positive edges, red dashed lines indicate negative edges. (b) 2-hop GraphSAGE~\cite{hamilton_inductive_2018} network based on the graph presented in (a). To get an embedding for node A, we traverse both positive and negative edges connected to A reaching nodes $h_A^1$. We repeat this step from nodes in $h_A^1$. Reached nodes are denoted as $h_A^0$. Values of $h_A^0$ equal to the position and orientation node features. Embeddings for nodes in $h_a^1$ are calculated by multiplying the mean-aggregated values of $h_A^0$ with a weight matrix shared across all nodes on this layer. To get an embedding for node $A$, another weight matrix is multiplied with the mean-aggregated feature embeddings of $h_A^1$. (c) Shallow MLP used for predicting existing/non-existent edges. Consists of two fully connected layers, with a ReLU activation being applied in-between them. }
    \vspace{-0.5cm}
    \label{fig:networks}
\end{figure*}

\section{GROWL}
Building upon the previous works on GNNs~\cite{hamilton_inductive_2018,zhang_link_2018}, in this paper we design a GNN based group detection approach, called GROWL (\textbf{GRO}up detection \textbf{W}ith \textbf{L}ink prediction). Similarly to~\cite{marco_cristani_social_2011, hedayati_reform_2020, hedayati_recognizing_2019, inaba_conversational_2016, setti_f-formation_2015, vascon_game-theoretic_2014,  zhang_beyond_2016, zhang_social_2018}, GROWL makes use of position and orientation features as input; however, differently from these methods, it determines whether a pair of people belong to the same group by taking into account their neighbourhood as well, rather than relying on pairwise affinity fields only. More explicitly, instead of using the deconstructed pairwise representation of a social scene as input, in GROWL people are treated as nodes in a graph. Groups are represented by sets of nodes connected by edges, which are learned by taking a holistic approach. 
GROWL is able to detect any number of groups (F-formations) of any cardinality (group size) in crowded scene images captured both from third-person view and robocentric (i.e., egocentric) view. As explained below, GROWL is comprised of two main steps: (A) graph representation generation and (B) link prediction. 

\subsection{Graph representation generation}
\label{sec:graphrep}
GROWL generates graph representations of social scenes by taking the position and orientation of people in the image. The position of people is not defined based on the camera angle and distance, instead it is a representation of where they are in a scene from a third-person view perspective. Since the resulting representation is an undirected graph, our solution is independent of group sizes and changes to the number of detected individuals.

Unlike SALSA~\cite{alameda-pineda_salsa_2016}, RICA~\cite{schmuck_rica_2020} is an egocentric (robocentric) dataset. To obtain graph representations, we applied a technique to map the position of the individuals captured from an egocentric view onto a to-down view (see Figure~\ref{fig:topdownrep}). To compute a top-down representation based on position, the scene can be defined as the area in front of a robot. Based on the robot's multimodal sensory inputs such as RGB, depth readings and distance information, individuals can be identified in front of the robot. To determine the position of the individuals (i.e., nodes) along the $x_{topdown}$ (horizontal) axis of the top-down representation, first the centroid of each individual's bounding box is located in the robocentric image, denoted by $c_k$, where \(k = \{1, ... , K\}\) and $K$ is the number of individuals in the scene. Then the $x$ coordinate of $c_k$ is normalised and assigned to $x_{topdown}$. For the $y_{topdown}$ coordinate of the representation, based on $c_k$ in the RGB input, the centroid is mapped onto the depth image to retrieve a single depth reading which can be normalised and assigned to $y_{topdown}$. Once both $x_{topdown}$ and $y_{topdown}$ values are computed, an individual (i.e., node) can be positioned in the top-down graph representation. This representation calculation is illustrated in Figure~\ref{fig:topdownrep}. 

To compute the orientation of an individual for the robocentric dataset, we used the Deep-orientation framework developed by Lewandowski et al~\cite{lewandowski_deep_2019}. Deep-orientation is a continuous regression based method, which builds on a biternion net architecture. Its architecture works best with depth images as input and produces an output of the estimated orientation in degrees. 
To estimate orientation of each individual from a robocentric view, we used depth images segmented based on the bounding boxes of people detected in their corresponding RGB images. 

Once position and orientation features for all individuals in a scene are computed, a graph can be generated based on the graph characteristics described in Section~\ref{sec:background_embedding}. This is done by setting node features to be the coordinates of people in a top-down scene representation and also adding the orientation of individuals as an additional feature. Edges between two nodes represent whether the individuals belong to the same interaction group, and following the work of Hedayati et al.~\cite{hedayati_reform_2020}, edge features are computed based on the effort angle and distance of two individuals. Effort angle is the amount in terms of radians that two individuals would need to turn to face each other, and distance is the Euclidean distance between two individuals.

\subsection{Link prediction in GROWL}
According to the findings of Zhang and Chen~\cite{zhang_link_2018}, selecting training samples for graph neural network based link predictors is non-trivial. Namely, a link predictor functions as a binary classifier that predicts whether an edge $E$ exists between two nodes $N$ in a graph $G = (N, E)$, where $E \subseteq N \times N$. However, as mentioned in Section~\ref{sec:background_link}, considering only existing edges (i.e., positive edges) in a graph during training leads to poor generalisation as non-existent edges (i.e., negative edges) will not be represented. Therefore, we perform negative injection by representing the created graphs both in terms of positive and negative edges, where 
positive edges are denoted by $E_p \subseteq E$, and edges created by negative sampling as $E_n \cap E = \oslash$. A representation of such a graph can be seen in Figure~\ref{fig:networks}~(a).

After performing negative injection, each of our graphs can be denoted by $G' = (N, E \cup E_n)$. In order to represent information of a node, we use a two-hop GraphSAGE~\cite{hamilton_inductive_2018} embedding with a mean aggregator. Looking at Figure~\ref{fig:networks}~(b), let A be a node from the graph created with negative injection in Figure~\ref{fig:networks}~(a). To calculate an embedding for node A, we traverse both positive and negative edges connected to A. Taking the nodes reached this way (collectively $h_A^1$), we traverse all edges once more from each of them, reaching nodes collectively denoted as $h_A^0$, making this a so-called two-hop embedding. The values of $h_A^0$ equal to the position and orientation node features. Embeddings for each node (e.g. $v$, $u$) in $h_A^1$ can be calculated by multiplying the mean-aggregated values of the previous ($h_A^0$) layer with a trainable weight matrix that is shared across all nodes on this layer, as proposed by Hamilton et al.~\cite{hamilton_inductive_2018}: \vspace{-0.2cm}

\begin{equation}
    \label{eq:embed}
    \textbf{h}_{A}^{1} \leftarrow \sigma \left ( \textbf{W}\cdot \textup{MEAN}\left ( \left \{ \textbf{h}_{A}^{0} \right \} \right ) \cup \left \{ \textbf{h}_{A}^{0},\forall u\in \emph{N}\left ( v \right ) \right \}\right ),
\end{equation}
where $\left \{ \textbf{h}_{A}^{0},\forall u\in \emph{N}\left ( v \right ) \right \}$ is the representation of the nodes in the neighbourhood of node $v$. Finally, to compute an embedding for node $A$, once again a different trainable weight matrix is multiplied with the mean-aggregated feature embeddings of layer $h_A^1$ .

During training, calculated embeddings of node pairs are concatenated and passed as inputs for a 2-layer Multi-layer Perceptron (MLP), which produces a label ($label \in [0, 1]$) for each edge. The MLP predictor consists of two fully connected layers, with a ReLU activation being applied in-between them, as shown in Figure \ref{fig:networks}~(c).

The created GraphSAGE embedding model and MLP predictor are trained end-to-end with the Adam optimizer and with binary cross-entropy loss for a number of epochs. To determine the optimal embedded feature size and number of epochs, we performed hyperparameter optimisation with the SALSA-PS training and test sets described in Section~\ref{sec:evalsetup}. 
We found that the best performing model was achieved with an embedded feature vector of size $20$ if trained for $100$ epochs. For a detailed description of the hyperparameter optimisation process, see Section~\ref{sec:paramopt}.

\begin{figure}[t]
  \centering
  \includegraphics[width=50mm]{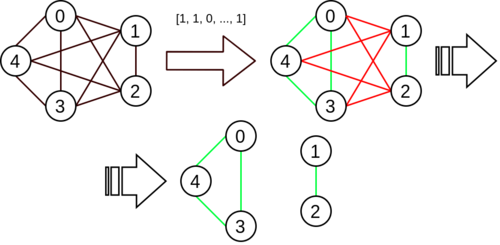}
  \caption{Edge elimination from a fully connected, non-directed graph based on GROWL's output. All edges of a graph are labelled by the MLP predictor, then the ones predicted as non-existent ($label=0$, red) are removed.}
  \vspace{-0.5cm}
  \label{fig:edge_elimination}
\end{figure}

Based on the labels calculated by the MLP predictor we can determine if two nodes are connected. However, we need to convert this information to a graph representation. To do so, we generate a fully connected, non-directed graph from all nodes and remove all edges which were assigned a label of $0$ as illustrated in Figure~\ref{fig:edge_elimination}, which results in a graph representation of detected groups. 

During testing, GROWL assumes a fully connected graph, and calculates node embeddings for each node via the trained 2-hop GraphSAGE model. The abovementioned MLP predictor can then use pairs of embedded node features to determine which edges are positive (existing) or negative (non-existent).

\section{EVALUATION}
\label{sec:eval}
We evaluated the proposed approach on both third-person view and egocentric view datasets and we compared our results with a baseline method based on Graph-Cuts (GCFF)~\cite{setti_f-formation_2015} and the current state-of-the-art interaction group detection method, REFORM~\cite{hedayati_reform_2020}.

\begin{table*}[t]
\caption{Comparison of GROWL against the state-of-the-art group detection methods, Graph-Cuts (GCFF) \cite{setti_f-formation_2015} and REFORM \cite{hedayati_reform_2020}, in terms of mean $F_1$-scores ($\overline{F_1}$) on the datasets SALSA \cite{alameda-pineda_salsa_2016} and RICA~\cite{schmuck_rica_2020}. SALSA-PS: SALSA Poster Session; SALSA-CPP: SALSA Cocktail Party; RICA-Y: GROWL tested on RICA with bounding boxes autimatically detected using YOLOv4~\cite{bochkovskiy_yolov4_2020}; GROWL-O: GROWL without orientation features; $\sigma$: standard deviation of $F_1$-scores, calculated over repeated evaluations. 
} 
\begin{center}
\begin{tabular}{|l||c|c||c|c||c|c||c|c||c|c|}
\hline
\textbf{\begin{tabular}[c]{@{}c@{}}Detection\\algorithm\end{tabular}} &
\textbf{\begin{tabular}[c]{@{}c@{}}SALSA\\$\overline{F_1}\uparrow$\end{tabular}} & 
\textbf{\begin{tabular}[c]{@{}c@{}}SALSA\\$\sigma\downarrow$\end{tabular}} & 
\textbf{\begin{tabular}[c]{@{}c@{}}SALSA\\-PS $\overline{F_1}\uparrow$\end{tabular}}&
\textbf{\begin{tabular}[c]{@{}c@{}}SALSA\\-PS $\sigma\downarrow$\end{tabular}}&
\textbf{\begin{tabular}[c]{@{}c@{}}SALSA\\-CPP $\overline{F_1}\uparrow$\end{tabular}}&
\textbf{\begin{tabular}[c]{@{}c@{}}SALSA\\-CPP $\sigma\downarrow$\end{tabular}}&
\textbf{\begin{tabular}[c]{@{}c@{}}RICA\\ $\overline{F_1}\uparrow$\end{tabular}} &
\textbf{\begin{tabular}[c]{@{}c@{}}RICA\\ $\sigma\downarrow$\end{tabular}} &

\textbf{\begin{tabular}[c]{@{}c@{}}RICA-Y\\ $\overline{F_1}\uparrow$\end{tabular}} &
\textbf{\begin{tabular}[c]{@{}c@{}}RICA-Y\\ $\sigma\downarrow$\end{tabular}}
\\
\hline \hline
GCFF & 59.7\% & 9.5 & 64.7\% & 12.4 & 59.3\% & 8.4 & 52.1\% & 30.1 & -- & -- 
\\
\hline
REFORM & -- & -- & 81.2\%\textsuperscript{*} & -- & -- & -- & -- & -- & -- & -- 
\\
\hline
GROWL-O & 22.5\% & 39.6& 33.7\% & 16.5 & 30.9\% & 41.5 & 44.5\% & 43.2 & 52.2\% & 43.7 
\\
\hline
GROWL & \textbf{84.4\%} & 22.5 & \textbf{91.4\%} & 11.1 & \textbf{84.1\%} & 24.1 & \textbf{84.7\%} & 27.4 & \textbf{74.7\%} & 41.9 
\\          
\hline
\multicolumn{5}{l}{\textsuperscript{*}\footnotesize{Taken from~\cite{hedayati_reform_2020}}}
\end{tabular}
\end{center}
\vspace{-0.9cm}
\label{tab:f1scores}
\end{table*}

\subsection{Group detection datasets}
To evaluate our proposed approach GROWL, we used the third-person view Synergetic sociAL Scene Analysis (SALSA) dataset~\cite{alameda-pineda_salsa_2016} and its subsets, Poster Session and Cocktail Party. In addition, we tested GROWL on a robocentric (egocentric) dataset called Robocentric Indoor Crowd Analysis (RICA)~\cite{schmuck_robocentric_2020}. 

The SALSA dataset~\cite{alameda-pineda_salsa_2016} contains third-person view recordings of social scenarios using $4$ RGB cameras. The annotated part of SALSA contains ground-truth information about $627$ and $500$ images from its Poster Session (SALSA-PS) and Cocktail Party (SALSA-CPP) subsets respectively. The majority of the recording comprises $18$ people interacting with each other for a duration of approximately $60$ minutes. The SALSA dataset is annotated with respect to position, pose and F-formation information within windows of $3$ seconds. 

The RICA dataset~\cite{schmuck_rica_2020} contains egocentric recordings of a social gathering from a mobile robot perspective. The dataset was recorded at around $10$ frames per second with the robot's on-board RGBD camera and comprises $40-50$ people interacting with each other for a duration of approximately 60 minutes. The RICA dataset contains manually annotated bounding box information of both individuals and interaction groups, which have been annotated frame by frame in the first over $8100$ images of the dataset. The annotated set holds information of $194$ continuous group occurrences, where group sizes range between $2-6$ people. RICA is considered challenging for a number of reasons. Due to the robocentic nature of the recordings, there is a significant noise stemming from motion blur, occlusion, and lighting condition changes compared to a stationary third-person view camera. Moreover, there are more participants and more static obstacles in the environment recorded in RICA compared to SALSA, leading to further noise factors.

\subsection{Experimental setup}
\label{sec:evalsetup}
To examine the generality of our solution, we trained and optimised GROWL on $60\%$ of our generated graphs randomly selected from the Poster Session subset of SALSA (SALSA-PS)~\cite{alameda-pineda_salsa_2016}. The optimised model is evaluated on the unseen test sets from SALSA and SALSA-PS as well as SALSA's Cocktail Party subset (SALSA-CPP) and RICA~\cite{schmuck_rica_2020}. Hedayati et al.~\cite{hedayati_reform_2020} followed a similar approach, namely training on $60\%$ of the SALSA-PS set and testing on the remaining $40\%$ of it.

We would like to highlight that the focus of our work is the detection of interaction groups. Detecting individuals and estimating their orientation in an end-to-end manner are beyond the scope of this paper. Therefore, for evaluating our method, we adopted the procedure same as the previous works by Setti et al.~\cite{setti_f-formation_2015} and Hedayati et al.~\cite{hedayati_reform_2020}, and we used the ground-truth information provided with the SALSA~\cite{alameda-pineda_salsa_2016} and RICA~\cite{schmuck_rica_2020} datasets for a fair comparison. More explicitly, we used the individuals' ground-truth position and orientation information as inputs when evaluating GROWL on the SALSA dataset~\cite{alameda-pineda_salsa_2016} and its subsets. In the case of the RICA dataset~\cite{schmuck_rica_2020}, we only used the ground-truth horizontal position of individuals ($x_{topdown}$) and calculated their vertical position ($y_{topdown}$) to achieve a top-down representation as described in Section~\ref{sec:graphrep}. Orientation was estimated from multimodal data using the method in \cite{lewandowski_deep_2019}, which relies on robocentric depth images (see Section \ref{sec:graphrep}). In addition to investigating to what degree an automatic human detector has impact on the accuracy of our pipeline, we fine-tuned YOLOv4~\cite{bochkovskiy_yolov4_2020} to detect people on the RICA dataset~\cite{schmuck_rica_2020}, and evaluated GROWL in two ways with automatically detected bounding boxes. The fine-tuned YOLOv4~\cite{bochkovskiy_yolov4_2020} had $55\%$ detection accuracy. Fine-tuning was performed on a randomly sampled $20\%$ of RICA~\cite{schmuck_rica_2020}, tuning out-of-the-box weights of YOLOv4~\cite{bochkovskiy_yolov4_2020} based on ground-truth human detections for $4000$ iterations. When testing the remaining set of RICA~\cite{schmuck_rica_2020}, and during the tests on GROWL, we considered detections successful if the intersection over union value of a detected bounding box compared to ground-truth human locations was above $40\%$.

We evaluated GROWL given the assumption that the automatically detected bounding boxes found by YOLOv4~\cite{bochkovskiy_yolov4_2020} represent all people in a scene. Due to this assumption, this evaluation (RICA-Y) calculated $F_1$-scores based on how accurately groups were detected by GROWL compared to ground-truth group annotations only including nodes representing automatically detected bounding boxes (i.e. excluding people from ground-truth groups who were not detected by YOLOv4). In reality, individuals who are not picked up by human detectors automatically result in False Negative node classifications when calculating group detection Precision and Recall scores, which consequently hinders the final $F_1$-score of the group detection. Therefore, we also evaluated GROWL with automatically detected bounding box inputs against complete ground-truth group annotations.

We compared GROWL with two methods, Graph-Cuts~\cite{setti_f-formation_2015} and REFORM~\cite{hedayati_reform_2020} as described below. Graph-Cuts (GCFF)~\cite{setti_f-formation_2015} is a benchmark used in most interaction group detection and F-formation detection studies. GCFF models the probability of each individual belonging to an F-formation as a Gaussian distribution. Then, by adding a Minimum Description Length (MDL) prior it builds a cost function to assign individuals to groups. Lastly, it adds a visibility constraint (stride) on the individual which is independent from the obtained probability. When choosing parameters for the GCFF~\cite{setti_f-formation_2015} benchmark, we chose the same parameters as reported by Hedayati et al.~\cite{hedayati_reform_2020}, $MDL = 30000$ and $stride = 0.7$ for a fair comparison. After selecting the MDL and stride values, GCFF can be run on the selected test sets.

REcognize F-FORmations with Machine learning (REFORM)~\cite{hedayati_reform_2020} uses pairs of individuals' position and orientation data. Similarly to GROWL, it predicts the link between two individuals using a binary classifier such as Weighted KNN, Bagged Tree or Logistic Regression, and then reconstructs groups from pairwise relationships through a greedy agreement algorithm. 

\subsection{Evaluation metrics}
\label{sec:evalmethods}

\begin{figure*}[t]
\centering
\tiny
\setlength\tabcolsep{3pt}
    \begin{tabular}{cccc}
        (a) & \includegraphics[width=0.42\textwidth]{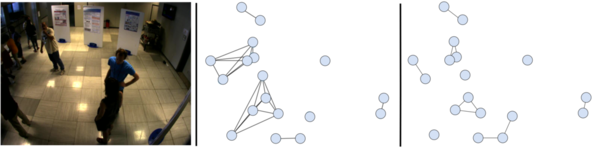}
        &
        (b) & \includegraphics[width=0.42\textwidth]{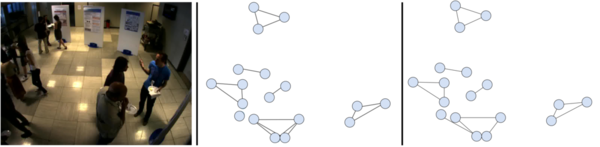}
        \\
        (c) & \includegraphics[width=0.42\textwidth]{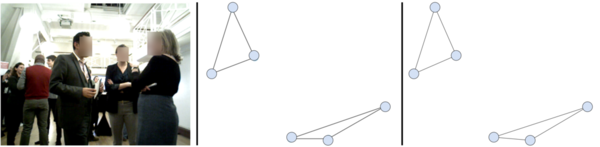}
        &
        (d) & \includegraphics[width=0.42\textwidth]{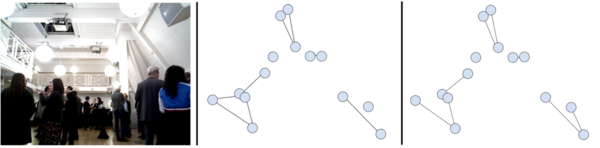}
        \\
    \end{tabular}
    \caption{This figure shows the RGB image (left), ground-truth (middle) and GROWL predicted (right) graph representations of a frame from the SALSA Cocktail Party dataset (a-b) and RICA dataset (c-d).}
    \vspace{-0.5cm}
    \label{fig:sample_graphs}
\end{figure*}

Calculating $F_1$-score has been the widely used metric to measure group detection performance. Setti et al.~\cite{setti_f-formation_2015} proposed a tolerance ratio ($T \in [0,1]$), where a group can be considered as correctly detected if at least $T$ of its members are correctly identified, and no more than $1-T$ individuals are incorrectly associated with it. In the previous work, $T$ has been commonly set to $T = 2/3$. Based on this ratio, an $F_1$-score can be obtained for each frame by calculating the precision and recall scores. 

\subsection{Parameter optimisation}
\label{sec:paramopt}
The goal of our hyperparameter optimisation is to find the optimal training length in epochs and the optimal feature embedding size of our model. We searched for parameters where the optimal embedded feature size was within a range between $2$ and $20$ and an optimal epoch length in the ranges $[10:50]$ with increments of $5$ and $[50:250]$ with increments of $50$. 
To determine which combination of parameters performs best, following the practice in our evaluation (see Section~\ref{sec:evalsetup}), after randomly shuffling their order we randomly selected $60\%$ of our generated graphs from the Poster Session subset of SALSA (SALSA-PS)~\cite{alameda-pineda_salsa_2016}. On this set, we performed a 10-fold cross-validation, resulting in approximately $362$ samples in each training set, and $38$ samples in validation sets. We measured the mean $F_1$-scores and their standard deviation for the validation folds. To ensure our results are not affected by a lucky training-test split, we repeated the above cross-validation procedure $3$ times, randomly selecting $60\%$ of the shuffled graphs from the SALSA-PS dataset each time, resulting in $30$ validation samples for each hyperparameter pair.

Based on the experiments, the best performing model was achieved with an embedded feature vector of size $20$ and an epoch count of $100$, which achieved a mean $F_1$-score of $\overline{F_1}=97.7\%$ with a standard deviation of $\sigma=13.1$.

\subsection{Experimental results}
\label{sec:expresults}
In Table~\ref{tab:f1scores}, we reported the average $F_1$-scores for predicted groups across test sets taken from the SALSA dataset~\cite{alameda-pineda_salsa_2016}, its subsets Poster Session (SALSA-PS) and Cocktail Party (SALSA-CPP)  
as well as RICA~\cite{schmuck_rica_2020}. We want to remind that GROWL was trained on a random, $60\%$ split of the SALSA-PS set only. To showcase that the resulting $F_1$-scores produced by GROWL were not a result of a lucky training set selection, we trained and evaluated GROWL $30$ times, randomly shuffling and selecting new training sets from SALSA-PS on each occasion. The results of our evaluation can be seen in Table~\ref{tab:f1scores}. Looking at the results, the generality of GROWL across different camera views is similar to that of GCFF~\cite{setti_f-formation_2015}; however, it performs significantly better, with an average of $20.8\%$ across all evaluated test sets. Moreover, GROWL outperforms the previous state-of-the-art method, REFORM~\cite{hedayati_reform_2020}, when evaluated on SALSA-PS by a margin of $10.2\%$ in terms of $F_1$-scores.

Examples of predicted interaction groups from the SALSA-CPP and RICA datasets can be seen in Figure~\ref{fig:sample_graphs}. We observe that GROWL is good at detecting separate, unconnected groups, however on occasion it connects two groups through a single node, producing both false positive and false negative group detections with a single mistake. 

As described in Section \ref{sec:evalsetup}, we also evaluated GROWL with automatically detected bounding boxes on the RICA dataset. Given the assumption that YOLOv4~\cite{bochkovskiy_yolov4_2020} could detect all individuals in a scene, GROWL achieved an average $F_1$-score of $74.7$. Without the abovementioned assumption, GROWL's average was $2\%$. These findings indicate that the reliable detection of individuals has a strong impact on the performance of group detection. While GROWL can identify interaction groups accurately, the inaccuracies of bounding box detection influence its performance significantly. In other words, GROWL's group detection accuracy is significantly hindered if YOLOv4~\cite{bochkovskiy_yolov4_2020} fails to detect people in the scene since undetected individuals introduce False Negative node classifications when calculating $F_1$-scores.

Based on our repeated evaluation, we observed that the standard deviation ($\sigma$) of measured average $F_1$-scores ($\overline{F_1}$) were high. This is due to the different randomly selected training sets and the resulting varying amount of positive and negative edges GROWL is trained on. According to our evaluation, this effect can be so radical that up to $2$ of the $30$ measurements will result in low ($\sim11\%$) mean $F_1$-scores when testing on SALSA, SALSA-CPP and RICA. These findings indicate that the single training and evaluation that Hedayati et al.~\cite{hedayati_reform_2020} reported is not sufficiently reliable, as the selection of training set has a significant impact on the performance of the group detection accuracy.

To investigate cases when the model produces low accuracies, we analysed the sample distribution it was trained on with different training samples. We observed that the number of positive edge samples ranges between $23200$ and $23850$ while the number of negative edge samples is between $88000$ and $90250$. Throughout different training sets the ratio between positive and negative samples is on average $20.8\%$ and $79.2\%$ respectively. However, when there are more than $90000$ negative samples, the measured mean accuracy drops significantly. Therefore, we want to highlight the importance of balancing the sampled training data, as a training set more skewed towards negative training samples hinders the accuracy of GROWL due to overfitting. Due to the nature of graphs and the relations between the nodes, this balancing is non-trivial and will be addressed in future work.

\section{Ablation studies}
In this section we report on how important the orientation feature is when detecting interaction groups. We also analyse the effect of calculating the orientation of people based on the recorded depth modality of a robocentric view. Lastly, we investigate the contribution of our applied negative injection.

\subsection{Contribution of orientation features}
While the SALSA dataset provides the ground-truth for orientation information, for the RICA dataset this information is not available and therefore as explained in Section~\ref{sec:graphrep} we used the Deep-orientation method for automatically estimating individuals' orientation from depth images. To demonstrate the reliability of orientation estimation and its contribution to the accuracy, we trained and tested GROWL through $30$ iterations as described in Section~\ref{sec:expresults}, using position information only for calculating the node embeddings, excluding the orientation. The resulting mean $F_1$-scores are provided in Table~\ref{tab:f1scores}.

We observed that the group detection accuracy dropped significantly without the orientation information. Based on these findings, we can hypothesise that GROWL creates an encoded representation similar to what Vascon et al.~\cite{vascon_game-theoretic_2015} described as attention frustum to identify interaction groups. Moreover, we can observe that even though the accuracy on the RICA dataset deteriorated significantly in case of both ground-truth and automatically detected inputs, compared to the resulting $F_1$-scores measured on the SALSA, SALSA-PS and SALSA-CPP test sets it performed better. This could be the result of noise introduced to raw node features during the graph representation calculation which translates the robocentric view to a third-person view.
This indicates that while the graph representation from robocentric view carries added noise, it still contributes significantly to the overall performance of the model when included as a feature.

\subsection{Contribution of negative injection}
According to the work of Zhang and Chen~\cite{zhang_link_2018}, we employed negative injection in order to achieve a model which can generalise better. In this section we verify that in the context of interaction group detection the exclusion of negative sampling results in overfitting, making the model prone to only predicting existing edges.

To investigate the contribution of training and evaluating with negative injection, as before, we trained and evaluated GROWL $30$ times as described in Section~\ref{sec:expresults} without generating training graphs via negative injection -- only taking graphs' existing (positive) edges. As expected, all evaluations concluded that the trained models achieve average $F_1$-scores of near 0\%. Upon further investigation, we confirmed that this is due to the model producing fully-connected graphs, which -- with our tolerance of $T=2/3$ used in our evaluation (see Section~\ref{sec:evalmethods}) -- results in scores of $0$ regarding both precision and recall scores. 

\section{CONCLUSION AND FUTURE WORK}
We proposed a Graph Neural Network (GNN) based approach called GROWL, to learn the links between individuals and their association with groups. GROWL learns the structural relationships between individuals through graphs by using individuals' position and orientation information to generate feature embeddings via GraphSAGE~\cite{hamilton_inductive_2018} and use them as node features. Pairs of these node features are given as inputs to a shallow MLP for predicting the existence of edges between the inspected nodes.

GROWL outperforms previous state-of-the-art approaches (GCFF~\cite{setti_f-formation_2015} and REFORM~\cite{hedayati_reform_2020}) while maintaining the same generality. Differently from the previous works~\cite{setti_f-formation_2015,hedayati_reform_2020}, we also demonstrated the viability of GROWL for detecting interaction groups in a fully automatic manner. Our ablation studies verified that orientation information is an important feature for interaction group detection. We also demonstrated that in case of robocentric images, calculating orientation from multimodal data is possible without significantly affecting group detection accuracy. Lastly, we verified that negative injection is crucial for training our model without overfitting, in line with the findings of Zhang and Chen~\cite{zhang_link_2018}. 
 
As a future work, we aim to optimise the sample representation in the selected training set to minimise overfitting further. We also plan to implement a method (e.g. majority voting strategy) to determine strong and weak edges to handle when two nodes -- which have multiple connections to their respective groups -- get connected to each other, producing a loosely connected, bigger group. Lastly, our results showed that the reliable detection of individuals is important for the subsequent task of group detection. Therefore, a promising path would be to extend GROWL into an end-to-end approach that can take an image of a scene as input and jointly perform person and group detection.  

\section*{ACKNOWLEDGMENTS}
We thank Toyota Motor Europe (TME) and Toyota Motor Corporation (TMC) for providing the Human Support Robot (HSR) as the robotic platform.
{\small
\bibliographystyle{ieee}
\bibliography{egbib}
}

\end{document}